\documentclass[letterpaper, 10 pt, conference]{IEEEtran}
\IEEEoverridecommandlockouts

\usepackage{graphicx}
\usepackage{pifont}
\usepackage{amsmath}
\usepackage{makecell}
\usepackage{amssymb}
\usepackage{amsfonts}
\usepackage{algorithmic}
\usepackage{array}
\usepackage{txfonts}
\usepackage{tabularx}
\usepackage{bbm}
\usepackage{bm}
\usepackage{textcomp}
\usepackage{stfloats}
\usepackage{url}
\usepackage{verbatim}
\usepackage[ruled,vlined,linesnumbered]{algorithm2e}
\usepackage{cite}
\usepackage{booktabs}
\usepackage{tikz}
\usepackage{color}
\usepackage{mathtools}
\usepackage[colorlinks=true, linkcolor=blue, citecolor=blue, urlcolor=blue]{hyperref}
\usepackage{rotating}
\usepackage{blkarray}
\usepackage{physics}
\usepackage{etoolbox}
\usepackage{cuted}
\usepackage{comment}
\usepackage{multirow}

\usepackage{caption}  

\usetikzlibrary{arrows}

\AfterEndEnvironment{strip}{\leavevmode}

\newcommand{\pname}{SemLoco}

\title{Watch Your Step: Learning Semantically-Guided Locomotion in Cluttered Environments}

\author{
        Denan Liang$^{1*}$,
        Yuan Zhu$^{1*}$,
        Ruimeng Liu$^1$,
        Thien-Minh Nguyen$^{1,2}$,~\IEEEmembership{Member,~IEEE}
        \\
        Shenghai Yuan$^1$,~\IEEEmembership{Member,~IEEE}
        Lihua Xie$^1$,~\IEEEmembership{Fellow,~IEEE}
\thanks{* indicates equal contribution.}
\thanks{This work is supported by the National Research Foundation of Singapore under its Medium-Sized Center for Advanced Robotics Technology Innovation.}
\thanks{$^1$School of Electrical and Electronic Engineering, Nanyang Technological University, 50 Nanyang Avenue, Singapore 639798, { \{shyuan,elhxie\}@ntu.edu.sg}.}
\thanks{$^2$School of Mechanical and Mining Engineering, The University of Queensland, St Lucia QLD 4072.}
}

\begin{document}



\twocolumn[{
\renewcommand\twocolumn[1][]{#1}
\maketitle
\begin{center}
    \centering
    \vspace{-0.5cm}
    \captionsetup{type=figure}
    \includegraphics[width=0.9\linewidth]{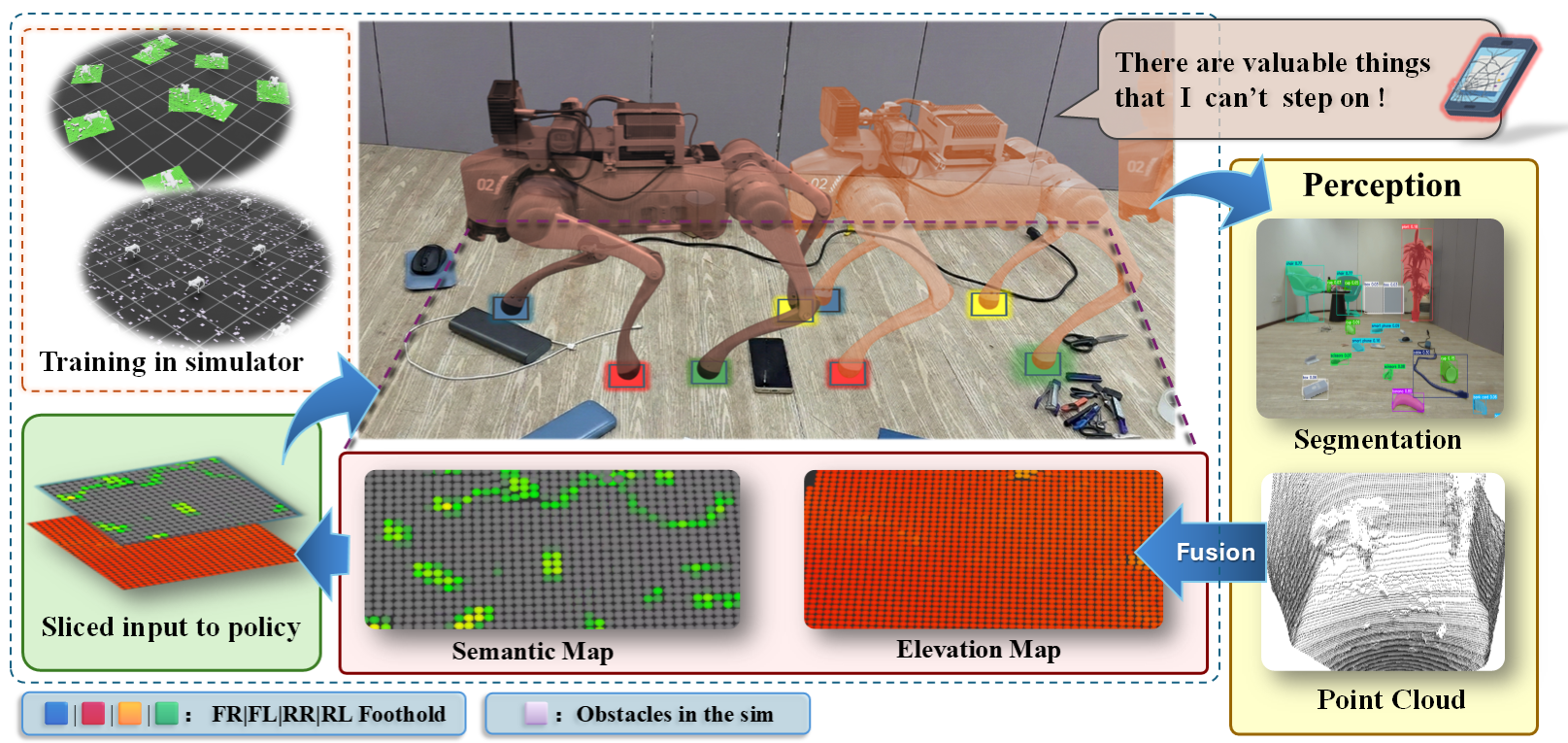}
    \caption{\textbf{{\pname}} Overview for semantic-aware locomotion in cluttered environments. This figure demonstrates a quadruped robot performing obstacle avoidance in a real-world scene with small sensitive objects. Compared to the traditional pure elevation map, {\pname} integrates semantic map to low-level control, enabling the controller to perform pixel-wise foothold safety inference. This allows for precise gait planning in both simulation and real-world experiments, preventing damage to valuable items.
    }
    \label{fig:overview}
\end{center}%
}]


\begin{abstract}
Although legged robots demonstrate impressive mobility on rough terrain, using them safely in cluttered environments remains a challenge. A key issue is their inability to avoid stepping on low-lying objects, such as high-cost small devices or cables on flat ground. This limitation arises from a disconnection between high-level semantic understanding and low-level control, combined with errors in elevation maps during real-world operation. 
To address this, we introduce {\pname}, a Reinforcement Learning (RL) framework designed to avoid obstacles precisely in densely cluttered environments. {\pname} uses a two-stage RL approach that combines both soft and hard constraints. It performs pixel-wise foothold safety inference, which enables more accurate foot placement.
Additionally, {\pname} integrates semantic map, allowing it to assign traversability costs instead of relying only on geometric data.
{\pname} greatly reduces collisions and improves safety around sensitive objects, enabling reliable navigation in situations where traditional controllers would likely cause damage. 
Experimental results further show that {\pname} can be effectively applied to more complex, unstructured real-world environments. A demo video can be viewed at \url{https://youtu.be/FSq-RSmIxOM}.

\end{abstract}

\bstctlcite{IEEEexample:BSTcontrol}
\vspace{-8pt}
\section{Introduction}
While quadruped robots have achieved exceptional success in robust locomotion and agile tracking across complex outdoor terrains \cite{hwangbo2019agile,miki2022perceptive,kumar2021rma}, their performance often degrades in densely cluttered indoor environments that demand slow, deliberate navigation. As their deployment transitions to indoor, human-centric spaces, a new critical requirement emerges.
In these environments, the primary objective shifts from merely maintaining the robot's dynamic balance to ensuring the safety of the surrounding environment. The core problem is that robots must execute high-precision foothold selection to avoid stepping on small, low-lying, or fragile objects (e.g., cables, small devices) scattered on the ground. It must proactively protect the human-centric environment through fine-grained contact point control.

\textbf{Existing} locomotion and navigation frameworks have made significant progress but fall short in this precise interaction. Current vision-based reinforcement learning (RL) policies and geometric planners effectively utilize exteroceptive perception to build elevation or occupancy maps for obstacle avoidance \cite{miki2022perceptive, rudin2022advancedskills, fankhauser2018terrain}. However, they lack explicit semantic prediction. Because they rely heavily on geometric representations, small objects (often only a few centimeters high) are frequently smoothed out as sensor noise or misclassified as traversable ground. Furthermore, while recent studies incorporate semantic data to enrich perception \cite{fan2022riskaware,miles2023terrainaware,yue2025safetypath}, they predominantly apply semantic constraints at the macro-level path or trajectory planning stage. They rarely pass these semantic preferences down to the joint-level control or explicit foothold selection.

The core \textbf{challenge} in cluttered, human-centric environments lies in bridging the gap between high-level semantic perception and high-frequency, low-level locomotion control. In dense scenarios, simply altering the global path is insufficient. Instead, the system must directly translate semantic properties into explicit foothold constraints, enabling the robot to precisely step over or around low-lying obstacles without compromising its dynamic stability.

To address this challenge, we propose the {\pname} framework, which directly embeds semantic information into low-level locomotion controller to achieve explicit semantic foothold prediction. Unlike traditional pipelines that decouple semantic navigation from geometric locomotion, our approach translates visual semantics into dynamic traversability costs that directly penalize unsafe foot placements. Powered by a two-stage RL training strategy—progressing from soft constraints with virtual objects to hard constraints with real rigid obstacles—{\pname} bypasses the ambiguities of geometric maps, ensuring precise and safe navigation in highly cluttered environments. In summary, the main contributions of this work are as follows:
\begin{itemize}
\item \textbf{{\pname} Framework:} We propose the {\pname} framework, which solves the exploration trap in the navigation of dense obstacles. This decouples spatial exploration from strict dynamics, enabling robust perception-motor mappings without early-stage kinematic collapse.
\item \textbf{Semantic-Geometry Decoupling:} We introduce a control paradigm that explicitly disentangles environmental semantics from geometric elevation. This resolves the inherent vulnerabilities of traditional elevation-based proxies, making the system highly robust against the depth sensor noise and specular reflections typical of small, fragile objects.

\item \textbf{Sim-to-Real Deployment and Validation:} We rigorously validate the proposed method across both simulations and real-world scenarios featuring fragile equipment. Experiments demonstrate that {\pname} drastically reduces the step collision rate, proving its robust autonomy and adaptability in cluttered environment.
\end{itemize}

\vspace{-8pt}
\section{Related Works}

\noindent\textbf{Locomotion Control:} Deep RL has become the mainstream paradigm for motion control of legged robots \cite{ha2025learning}. In recent years, researchers have no longer focused solely on walking stability but have gradually taken obstacle avoidance capability in environmental interaction as a core objective.
In precise obstacle avoidance, foothold planning is a key component. Early optimization-based methods (e.g., model predictive control) were limited by computational efficiency and map accuracy, making it difficult to handle environments with dense obstacles \cite{grandia2023perceptive,kim2019highly}.
Recently, Rudin et al. proposed using RL to learn various gaits in simulation and demonstrated obstacle avoidance capabilities in environments with scattered obstacles \cite{9981198}, but their strategy still relies on elevation maps and has limited recognition ability for low-lying objects.
To address this issue, Miki et al. proposed mapping perceptual information directly to foot trajectories through end-to-end learning, achieving real-time obstacle avoidance in complex terrains \cite{miki2022perceptive}, but this method still lacks semantic understanding of obstacles.

For the accuracy of footholds, Jenelten et al. further proposed combining RL with whole-body control and introducing a differentiable cost function to achieve high-precision avoidance \cite{jenelten2020perceptive}. However, this method decouples foothold search from gait phase and the robot's current motion, making it prone to generating incoherent footholds in dense obstacle environments and falling into local optima. {\pname} \ redesigns the foothold calculation method by combining the Raibert heuristic with dynamic grid search, directly integrating gait phase and motion commands. This tightly couples foothold adjustment with the robot's dynamic behavior, significantly improving precise avoidance performance in scenarios with dense low-lying obstacles.

\noindent\textbf{Semantic Perception in Robot Navigation:} Beyond purely geometry-based navigation, recent works leverage semantic information to enhance planning. Several modular approaches employ semantic segmentation models to obtain object masks and assign heuristic costs to construct dense semantic maps \cite{erni2023mem,maturana2017real} or semantic graphs \cite{kremer2023s}. \cite{kim2024learning} learns a semantic traversability estimator through an automated annotation pipeline on egocentric videos, simplifying the derivation of traversal cost maps. Furthermore, semantic-aware planners are built upon these representations to incorporate object-level risk into navigation decisions \cite{cai2022risk,achat2022path}. More recently, \cite{roth2024viplanner} proposes an end-to-end path planning framework that encodes semantic information in the latent space, improving planning efficiency. \cite{yang2023corl_semanticsloco} integrates semantic semantics into quadrupedal navigation, yet its impact is confined to the tuning of gait parameters, and velocity levels. \cite{aegidius2025watch} generates a passability cost map for local path planning by extracting rich image semantic features through the vision transformer DINOv2. 
These semantic-aware methods primarily operate at the high-level path planning stage and do not explicitly reason about foothold selection for legged robots. In cluttered and narrow environments, even if fragile or undesirable objects are recognized by the high-level planner, the absence of foothold-level constraints may still cause the robot to step on them, limiting the effectiveness of semantic awareness in contact-rich locomotion. To address the above issues, {\pname} \, for the first time, combines semantic information with foothold planning for legged robots, using it to assign traversability costs to different types of obstacles, thereby establishing an effective connection between high-level semantic understanding and low-level motion control.

\section{Methodology}

\begin{figure*}[t]
    \vspace{-10pt}
    \centering
    \includegraphics[width=1.0\linewidth]{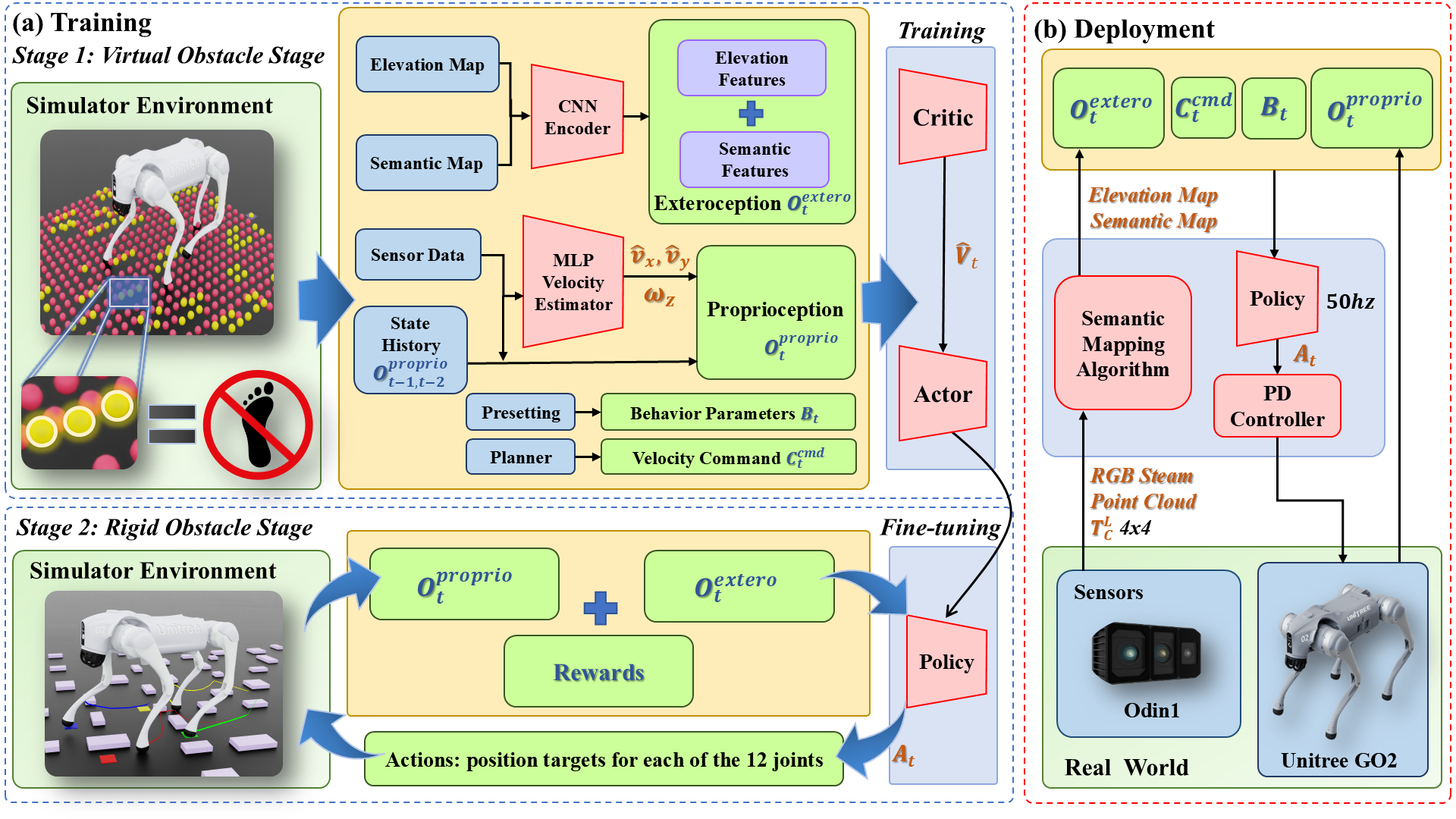}
    \caption{
    Framework of \textbf{\pname}.     
    Sub-modules have different styles based on their functions. Among them, the red trapezoid represents the neural network, the blue rectangle represents unprocessed raw data, and the green rectangle represents processed ready-to-use data.
    (a) \textbf{Training in the simulator:}
    In stage 1, we use virtual obstacles (Highly yellow spheres). Although the robot walks on flat ground, it receives a virtual perception map containing height and semantic information to simulate scenarios with real obstacles.
    In stage 2, we use rigid obstacles. The robot receives corresponding real perception information, and fine-tunes the policy to improve task performance.
    (b) \textbf{Deployment in the real world:}
    Exteroception information is obtained by Odin1 and fed into the semantic graph algorithm.
    }
    \label{fig:framework}
    \vspace{-10pt}
\end{figure*}


\subsection{Reinforcement Learning}

\subsubsection{\textbf{Observation Space}}

The Symmetric Actor-Critic (A2C) algorithm is adopted as the RL framework. The policy observations, $\boldsymbol{O}_{t} \in \mathbb{R}^{1513}$, are defined as:
\begin{equation}
    \label{eq:O}
    \boldsymbol{O}_t = \left[\boldsymbol{C}_t^{\text{cmd}}, \boldsymbol{B}_t, \boldsymbol{O}_t^{\text{proprio}}, \boldsymbol{O}_t^{\text{extero}} \right],
\end{equation}
where $ \boldsymbol{C}_{t}^{\text{cmd}} \in \mathbb{R}^{3}$ are the desired velocity commands, including the base linear velocities $\boldsymbol{v}_{x}^{\text{cmd}} \in \mathbb{R}^{1}$ in the x-axes, $\boldsymbol{v}_{y}^{\text{cmd}} \in \mathbb{R}^{1}$ in the y-axes and the base angular velocity $\boldsymbol{\omega}_{z}^{\text{cmd}} \in \mathbb{R}^{1}$ in the yaw axis. 

The $\boldsymbol{B}_t \in \mathbb{R}^{13}$ specify the behavior parameters used for generating stable four-beat quadrupedal contact pattern(walking), which is inspired by \cite{margolis2023walk,siekmann2021sim}. They are defined as:
\begin{equation}
    \label{eq:b}
\boldsymbol{B}_t = \left[\boldsymbol{h}_{\text{base}}, \boldsymbol{s}_{\text{feet}}, \boldsymbol{\theta}_{1-3}, \boldsymbol{t}_{\text{feet}}, \boldsymbol{f}, \boldsymbol{d}, \boldsymbol{w}, \boldsymbol{l} \right],
\end{equation} where $\boldsymbol{h}_{\text{base}} \in \mathbb{R}^{1}$, $\boldsymbol{s}_{\text{feet}} \in \mathbb{R}^{1}$, ${f} \in \mathbb{R}^{1}$ represent the height of robot’s base, the footswing height and the frequency of contact. $\boldsymbol{\theta}_{1-3} \in \mathbb{R}^{3}$, $\boldsymbol{t}_{\text{feet}} \in \mathbb{R}^{4}$, ${d} \in \mathbb{R}^{1}$, $\boldsymbol{w} \in \mathbb{R}^{1}$, $\boldsymbol{l} \in \mathbb{R}^{1}$ denote the timing offsets for three pairs of feet, the contact state timer for each foot, the duty, the stance width and length.

The $\boldsymbol{O}_t^{\text{proprio}} \in \mathbb{R}^{57}$ denote proprioception measured from IMU, joint encoders and velocity estimator: 
\begin{equation}
    \label{eq:O2}
    \boldsymbol{O}_t^{\text{proprio}} = \left[\boldsymbol{\hat{v}}_{t}, \boldsymbol\omega_{t}, \boldsymbol{g}_{t}, \boldsymbol{q}_{t}, \boldsymbol{\dot{q}}_{t}, \boldsymbol{a}_{t-1}, \boldsymbol{a}_{t-2}\right],
\end{equation}
where, $\boldsymbol{\hat{v}}_{t} \in \mathbb{R}^{3}$ denote the real base linear velocity, which is estimated by the velocity estimator. The $\boldsymbol{\omega}_{t} \in \mathbb{R}^{3}$, $\boldsymbol{g}_{t} \in \mathbb{R}^{3}$, $\boldsymbol{q}_{t} \in \mathbb{R}^{12}$ and $\boldsymbol{\dot{q}}_{t} \in \mathbb{R}^{12}$ denote the real base angular velocity, gravity vector, joint angle and joint angular velocity, which are measured from the joint encoder and IMU. The $\boldsymbol{{a}}_{t-1} \in \mathbb{R}^{12}$ and $\boldsymbol{{a}}_{t-2} \in \mathbb{R}^{12}$ are the previous actions of the first two steps. 

The $\boldsymbol{O}_t^{\text{extero}} \in \mathbb{R}^{1440}$ are defined as information about exteroception. Environmental information includes the elevation map $\boldsymbol{H}_{t}^{\text{elev}}\in \mathbb{R}^{720}$ and the semantic map $\boldsymbol{H}_{t}^{\text{sem}} \in \mathbb{R}^{720}$, both of which are centered at the center of gravity of the robot. Both maps are defined over a region that spans 1.5 m in the x-direction and 1.2 m in the y-direction. At 0.05m resolution, this results in a 30 $\times$ 24 grid of sampled cells.
\subsubsection{\textbf{Action Space}}
The action $\mathbf{a}_{t} \in \mathbb{R}^{12}$, are defined as the differences between the nominal position and the target position for each of the twelve joints of the robot, which are outputted by the actor network. A proportional derivative (PD) controller is used to track desired joint positions by converting positions to torques. The proportional gain ${k}_{p}$ and the derivative gain ${k}_{d}$ are 20 and 0.5 respectively.
\subsubsection{\textbf{Network Architecture}}
The framework is illustrated as Fig.~\ref{fig:framework}. Our policy consists of two Multilayer Perceptrons (MLPs) for the actor network and the critic network respectively, a base velocity estimator and a Convolutional Neural Network (CNN) encoder. 
The CNN encoder extracts height features from elevation map and semantic features from semantic map. The feature vector extracted by the CNN module is concatenated with all other proprioceptive observations, and the concatenated vector is then fed into an MLP with hidden layer sizes of [512, 256, 128] and ELU activation function. The base velocity estimator, which is a MLP with hidden layer sizes [256, 128] and ELU activations, is trained in a supervised manner.
\subsection{Semantic-Aware Adaptive Foothold Planning}
We proposed a semantic-aware Raibert Heuristic to overcome the vulnerability of traditional controllers in cluttered environments. Rather than relying solely on the neural network for implicit obstacle avoidance, our approach explicitly integrates local obstacle constraints into the kinematic planning loop. By dynamically evaluating semantic collision costs over a localized search grid, this method refines the nominal, blind footholds into intrinsically safe targets before they are tracked by the RL policy.

The standard Raibert Heuristic calculates the ideal foothold for leg $\boldsymbol{i}$ based on the robot’s base coordinate system to maintain velocity tracking and offset the effects of inertia and angular velocity.\cite{raibert1986legged, kim2019highly} The nominal stance position is defined as $\boldsymbol{P}_{nom}^{i} = [\boldsymbol{x}_{\text{nom}}^{i}, \boldsymbol{y}_{\text{nom}}^{i}]^T$ based on the desired stance width $\boldsymbol{w}$ and length $\boldsymbol{l}$. Let $T_{\text{stance}}$ be the stance duration, which is derived from the desired gait frequency $\boldsymbol{f}$ and duty factor $\boldsymbol{d}$ as $\boldsymbol{T}_{\text{stance}} = \frac{\boldsymbol{d}}{\boldsymbol{f}}$. The nominal Raibert position $P_{\text{raibert}}^{i}$ in the body frame is calculated by incorporating the desired linear velocity $\boldsymbol{v}_{x}^{\text{cmd}}$, $\boldsymbol{v}_{y}^{\text{cmd}}$ and the yaw angular velocity $\boldsymbol{\omega}_{z}^{\text{cmd}}$:
\begin{equation}
    \label{eq:raibert1}
\boldsymbol{P}_{\text{raibert}}^{i} = \boldsymbol{P}_{\text{nom}}^{i} + \Delta \boldsymbol{P}_{\text{lin}}^{i} + \Delta \boldsymbol{P}_{\text{yaw}}^{i},
\end{equation}
where the translational offset $\Delta \boldsymbol{P}_{\text{lin}}^{i}$ and the rotational offset $\Delta \boldsymbol{P}_{\text{yaw}}^{i}$ are defined as:
\begin{equation}
    \label{eq:raibert2}
\Delta \boldsymbol{P}_{\text{lin}}^{i} = \begin{bmatrix} \frac{\boldsymbol{T}_{\text{stance}}}{2} \boldsymbol{v}_{x}^{\text{cmd}} \\ 0 \end{bmatrix},
\end{equation}

\begin{equation}
    \label{eq:raibert3}
\Delta \boldsymbol{P}_{\text{yaw}}^{i} = \begin{bmatrix} 0 \\ \frac{\boldsymbol{T}_{\text{stance}}}{2} \boldsymbol{\omega}_{z}^{\text{cmd}} \cdot \boldsymbol{x}_{\text{nom}}^{i} \end{bmatrix},
\end{equation}
where the $\boldsymbol{x}_{\text{nom}}^{i}$ encodes the longitudinal position of the leg (positive for front legs, negative for rear legs), thereby automatically assigning the correct lateral stepping direction induced by the body's yaw rotation.

Building upon this kinematic prior, our proposed method performs a localized optimal search around the nominal Raibert position $\boldsymbol{P}_{\text{raibert}}^{i}$ leveraging the semantic grid map $\boldsymbol{H}_{t}^{\text{sem}}$. We construct a localized $M \times M$ search grid $\mathcal{G}^{i}$ with resolution $\Delta_{\text{grid}}$ centered around $\boldsymbol{P}_{\text{raibert}}^{i}$. For each candidate point $\boldsymbol{p}_{k} \in \mathcal{G}^{i}$, we transform it to the world frame and check for virtual collisions against known obstacles from the semantic module. Crucially, rather than treating all spatial deviations equally, quadrupeds exhibit strong kinematic preferences during obstacle avoidance: shortening the stride (stepping backward relative to the nominal point) is biomechanically safer than over-stretching forward, while lateral deviations severely compromise the support polygon and should be strictly avoided. To capture this, we design a kinematically informed cost function $J(\boldsymbol{p}_k)$ composed of an asymmetric directional deviation cost and a collision penalty:
\begin{equation}
    \label{eq:raibert5}
J(\boldsymbol{p}_k) = C_{\text{dir}}(\boldsymbol{p}_k) + w_{\text{col}} \cdot \mathbbm{1}_{\text{col}}(\boldsymbol{p}_k),
\end{equation}
where the directional deviation cost $C_{\text{dir}}(\boldsymbol{p}_k)$ is defined in the robot's local heading frame with longitudinal offset $\Delta x$ and lateral offset $\Delta y$:
\begin{equation}
    \label{eq:raibert6}
C_{\text{dir}}(\boldsymbol{p}_k) = w_{x}^{+} \max(\Delta x, 0) + w_{x}^{-} \max(-\Delta x, 0) + w_{y} |\Delta y|.
\end{equation}
To enforce the aforementioned kinematic priors, we strictly set the penalty weights as $w_{y} \gg w_{x}^{+} > w_{x}^{-}$. The collision penalty, $\mathbbm{1}_{\text{col}}(\boldsymbol{p}_k)$, is an indicator function that evaluates to $1$ if the candidate point $\boldsymbol{p}_k$ triggers an Axis-Aligned Bounding Box (AABB) collision with any semantic obstacle, whose boundaries are uniformly dilated by the foot radius $r_{\text{foot}}$ to ensure a strict safety margin, and $0$ otherwise. We assign a highly punitive weight $w_{\text{col}} \gg w_{y}$ to ensure that unsafe locations are fundamentally vetoed.The final optimized target foothold $\boldsymbol{P}_{\text{target}}^{i}$ is derived by minimizing the cost function across all candidates:

\begin{equation}
    \label{eq:raibert4}
\boldsymbol{P}_{\text{target}}^{i} = \arg\min_{\boldsymbol{p}_k \in \mathcal{G}^{i}} J(\boldsymbol{p}_k).
\end{equation}

In this task, the semantic information is defined as "fragility"(from 0 to 1). To obtain semantic information, we construct a semantic traversability map by projecting semantic predictions from RGB images into a global grid map. At time $t$, an open-vocabulary semantic segmentation model \cite{sun2025yolo} predicts a class probability distribution $P(S_t=c|u,v)$ for each pixel $(u,v)$ in the RGB image. Each pixel is associated with a 3D point from the aligned point cloud and projected to a grid cell $g$ in the elevation map \cite{fankhauser2018terrain}. For each cell, we maintain a probability distribution over semantic classes $P_t(c|g)$, which is updated using Bayesian fusion:
\begin{equation}
    \label{eq:semantic}
    P_t(c|g) \propto P(S_t=c|p_i)P_{t-1}(c|g),
\end{equation}
where $p_i$ is the observed point projected to cell $g$. Each class $c$ is assigned a predefined traversal cost $\phi(c)$, and the final traversability cost of the cell is computed as the expected cost:
\begin{equation}
    \label{eq:semantic2}
    M(g)=\sum_{c\in\mathcal{C}} P_t(c|g)\phi(c).
\end{equation}

\subsection{Reward Functions}
To facilitate the learning of highly dynamic yet safe locomotion, we adopt a multiplicative reward structure from\cite{ji2022concurrent}. The total reward $R_{total}$ is defined as the product of primary task rewards and an exponential auxiliary penalty multiplier:
\begin{equation}
    \label{eq:total_reward}
    R_{total} = r_{\text{primary}} \cdot \exp\left(c_{\text{penalty}} r_{\text{penalty}}\right),
\end{equation}
where $r_{\text{task}} \geq 0$ encourages the completion of primary objectives, $r_{\text{aux}} \leq 0$ penalizes undesirable behaviors, and $c_{\text{aux}}$ is a positive scaling factor. This form largely resolves the trade-off issue between different rewards. The terms in \eqref{eq:total_reward} are further elaborated below.

\subsubsection{\textbf{Primary Objective Rewards} (\texorpdfstring{$r_{\text{primary}}$}{rprimary})}
In highly cluttered environments, tracking the velocity command is insufficient for survival, and precise foot placement is equally important. Thus, our primary reward integrates both kinematic velocity tracking and semantic-aware foothold tracking:
\begin{equation}
    \label{eq:task_reward}
    r_{\text{primary}} = w_{\text{vel}} \cdot r_{\text{vel}} + w_{\text{sem}} \cdot r_{\text{semantic}},
\end{equation}
where the velocity tracking term $r_{\text{vel}}$ and the semantic-aware foothold tracking term are defined as:
\begin{equation}
    \label{eq:vel_reward}
    r_{\text{vel}} = \exp\left(-\frac{\|\boldsymbol{v}_{xy} - \boldsymbol{v}_{xy}^{\text{cmd}}\|_2^2}{\sigma_{v}}\right) + \exp\left(-\frac{(\boldsymbol{\omega}_{z} - \boldsymbol{\omega}_{z}^{\text{cmd}})^2}{\sigma_{\omega}}\right),
\end{equation}

\begin{equation}
    \label{eq:sem_reward}
    r_{\text{sem}} = \sum_{i=1}^{4} \mathbbm{1}_{\text{swing}}^{i} \cdot \exp\left(-\frac{\|\boldsymbol{p}_{\text{foot}}^{i} - \boldsymbol{P}_{\text{target}}^{i}\|_2^2}{\sigma_{\text{foot}}}\right),
\end{equation}
where $\mathbbm{1}_{\text{swing}}^{i} \in \{0, 1\}$ is the swing phase indicator.
The semantic tracking term $r_{\text{sem}}$ enforces the policy to accurately step on the optimal, collision-free targets $\boldsymbol{P}_{\text{target}}^{i}$ generated by our explicit spatial planner. By formulating this as a dense positive tracking reward during the swing phase, we provide a continuous gradient for the policy.

\subsubsection{\textbf{State and Dynamic Constraint Penalties} (\texorpdfstring{$r_{\text{penalty}}$}{rpenalty})}

The penalty term, $r_{\text{penalty}} = \sum w_k c_k$, aggregates various negative constraints to bridge the reality gap, maintain body balance, and optimize energy efficiency. To ensure robust obstacle traversal, we design a minimum foot clearance penalty rather than a strict trajectory tracking constraint. This encourages the policy to lift its feet sufficiently high to avoid toe-stubbing, without penalizing it for stepping higher than the reference when navigating severe cluttered obstacles. 

\begin{figure}[htbp]
    \vspace{-5pt}
    \centering
    \includegraphics[width=0.95\linewidth]{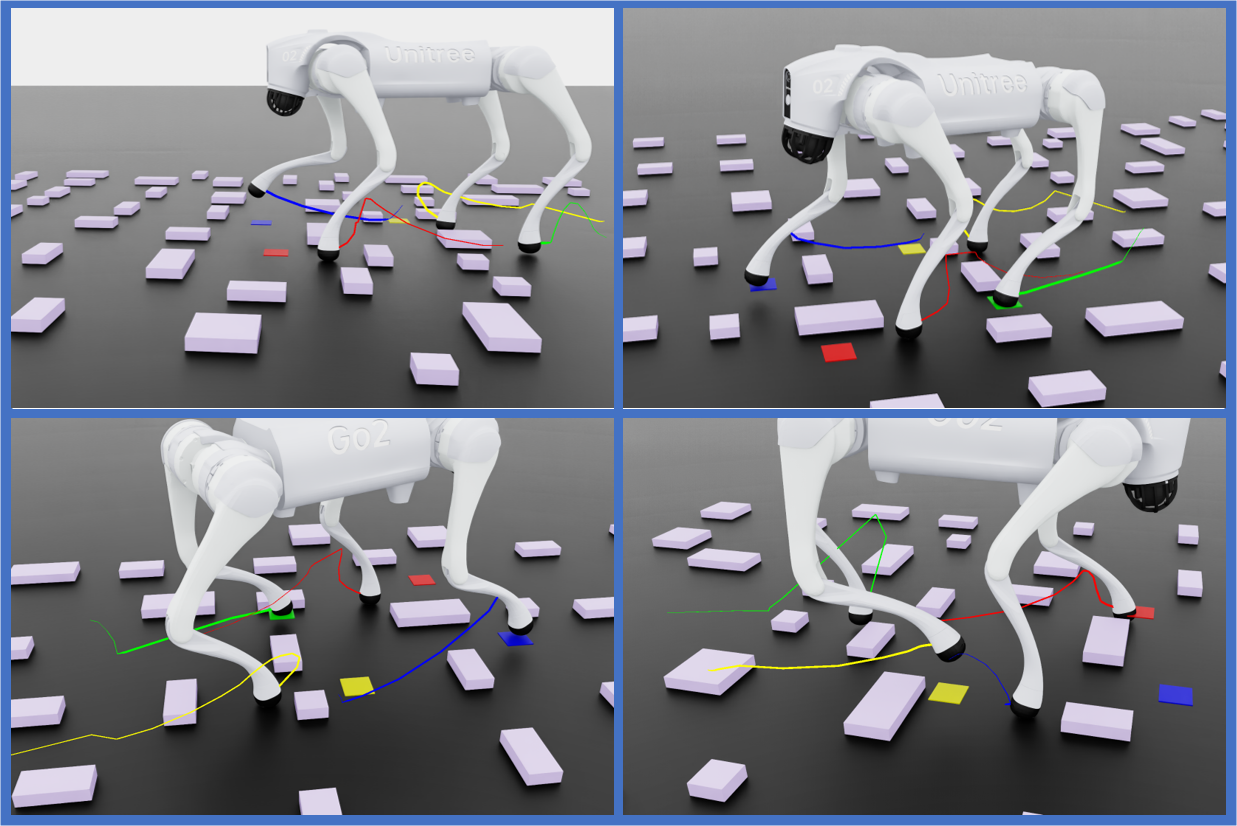}
    \caption{
        Footswing tracking for each leg.
    }
    \label{fig:footswing}
    \vspace{-5pt}
\end{figure}

During the swing phase, the normalized swing progress $\phi_{\text{swing}}^{i} \in [0, 1]$ for leg $i$ is calculated based on the stance duty factor. To synthesize an asymmetric safety-first swing profile characterized by rapid liftoff and a prolonged airborne plateau (Fig.~\ref{fig:footswing}), we formulate the time-varying reference height $z_{\text{ref}}^{i}$ using a square-root sine function:
\begin{equation}
    \label{eq:swing_ref}
    z_{\text{ref}}^{i} = \boldsymbol{s}_{\text{feet}} \cdot \sqrt{\sin(\pi \phi_{\text{swing}}^{i})} + \delta_z,
\end{equation}
where $\boldsymbol{s}_{\text{feet}}$ is the footswing height and $\delta_z$ is a fixed safety margin ($\delta_z = 0.02$ m). 

To enforce this as a lower-bound safety threshold, we utilize a ReLU function, mathematically expressed as $\max(0, x)$, to selectively penalize the foot only when its actual vertical position $z_{\text{footswing}}^{i}$ falls below the reference curve. The clearance penalty is defined as:
\begin{equation}
    \label{eq:clearance_penalty}
    c_{\text{clearance}} = \sum_{i=1}^{4} \mathbbm{1}_{\text{swing}}^{i} \cdot \max(0, z_{\text{ref}}^{i} - z_{\text{footswing}}^{i})^2,
\end{equation}
where the indicator function $\mathbbm{1}_{\text{swing}}^{i} \in \{0, 1\}$ ensures the penalty is solely evaluated during the swing phase.

Furthermore, dynamic constraints are applied to strictly restrict large joint torques and velocities. Finally, severe collision penalties are imposed on non-foot rigid bodies (e.g., thighs and base) to guarantee physical safety.

\subsection{Learning Semantic-Aware Locomotion via Two-Stage RL}
Learning semantic-aware locomotion from scratch in a highly cluttered environment poses a significant exploration challenge. Direct training with rigid physical obstacles frequently causes a large number of penalties, severely restricting the agent's ability to collect meaningful reward signals and leading to overly conservative policies. Consequently, the policy often converges to suboptimal local minima, failing to master either stable locomotion or reliable obstacle avoidance. To address this, we design a two-stage RL approach.

\subsubsection{\textbf{Virtual Obstacle Stage}}
In the first stage, we create a virtual training environment where obstacles are visually and semantically represented in $\boldsymbol{H}_{t}^{\text{elev}}$ and $\boldsymbol{H}_{t}^{\text{sem}}$, but physical collisions are disabled, as shown in Fig.~\ref{fig:framework}(a). The robot's limbs can pass through obstacles without triggering physical reaction forces or ending the episode. This means the robot will neither trip over obstacles nor push them away. During this stage, the policy relies solely on the Semantic-Aware Raibert Heuristic Reward ($r_{\text{sem}}$) to adjust its foot placement. Working together with the curriculum, this safe stage allows the policy to gradually progress from basic stable locomotion to semantic-aware foot placement. This helps the robot build a robust mapping from visual inputs to spatial leg-swing coordination, without the severe penalties of physical collisions.

\subsubsection{\textbf{Rigid Obstacle Stage}}
After the policy converges in the soft dynamic stage, we transfer it to the second stage, where full physical interactions are enabled. Obstacles are created as rigid bodies, and their density and variety are increased compared to the first stage, further improving the robot's obstacle avoidance ability. In this stage, the policy adjusts the previously learned motion patterns to handle complex physical interactions, such as friction.

\subsection{Curriculum Learning for Generalization and Robustness}
We design two types of curriculum:Velocity Curriculum and Obstacle Density Curriculum.
\subsubsection{\textbf{Velocity Curriculum}}
We use a grid-based adaptive curriculum form \cite{margolis2024rapid}. The linear velocities and angular velocities are uniformly discretized into a multi-dimensional grid of difficulty bins. During training, the sampling probabilities of these bins are dynamically updated based on the agent's historical tracking rewards. In addition to velocity commands, training can be conditioned on behavior parameters. While this approach did not significantly benefit our primary obstacle-avoidance task, expanding the instruction space remains an effective strategy for versatile terrain adaptation.
\subsubsection{\textbf{Obstacle Density Curriculum}}
\begin{figure}[htbp]   
    \vspace{-10pt}
    \centering    
    
    \includegraphics[width=0.32\linewidth]{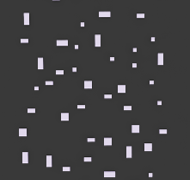}   
    \hfill 
    \includegraphics[width=0.32\linewidth]{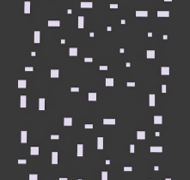}   
    \hfill 
    \includegraphics[width=0.32\linewidth]{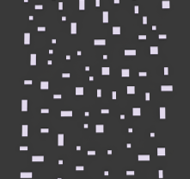}      

    \caption{Examples of obstacle curricula.}        
    \label{fig:Terrain_Curriculum}
    \vspace{-13pt}
\end{figure}

If the robot starts training in an environment with too many obstacles, it may often fail and receive high penalties, leading to a conservative strategy where it just stays in place. To avoid this, we use an obstacle density curriculum to control the difficulty of the training environment. We use $\rho_{\text{obst}}$ to represent the density of obstacles. As the robot gets better at tracking velocity commands, we gradually increase $\rho_{\text{obst}}$ until it reaches the maximum level. Fig.~\ref{fig:Terrain_Curriculum} provides examples.

\subsection{Training Setup in Simulator and Real-world Deployment}
\textbf{Training:} We conduct two-stage training environments in the IsaacLab framework \cite{mittal2025isaac} and use Proximal Policy Optimization \cite{schulman2017proximal} to train policies. Each stage is trained with 4096 parallel environments of Unitree Go2 robot. We train on a single NVIDIA RTX 5090 GPU, with the soft constraint stage taking 1.5 hours and the hard 3.5 hours.

\textbf{Real-World Deployment:} We deploy SemLoco on a Unitree Go2 equipped with a head-mounted RGB-ToF module Odin1 and an AGX Orin. To preserve the robot's nominal dynamics, all computations—including semantic segmentation, map fusion, and policy inference—are executed locally. The control policy runs at 50 Hz, while the asynchronous perception front-end updates at 10 Hz.

\section{Experiments}

    
\begin{table*}[t]
\centering
\caption{Ablation experiment under different obstacle densities.}
\label{tab:quantitative_results}
\resizebox{\linewidth}{!}{ 
\begin{tabular}{lcccccccccccc}
\hline
\toprule
\multirow{2}{*}{\textbf{Policy}} & \multicolumn{3}{c}{\textbf{10 obstacles $\mathrm{m}^{-2}$}} & \multicolumn{3}{c}{\textbf{15 obstacles $\mathrm{m}^{-2}$}} & \multicolumn{3}{c}{\textbf{20 obstacles $\mathrm{m}^{-2}$}} & \multicolumn{3}{c}{\textbf{25 obstacles $\mathrm{m}^{-2}$}} \\
\cmidrule(lr){2-4} \cmidrule(lr){5-7} \cmidrule(lr){8-10} \cmidrule(lr){11-13}
& $D \uparrow$ & $S \uparrow$ & $C \downarrow$ & $D \uparrow$ & $S \uparrow$ & $C \downarrow$ & $D \uparrow$ & $S \uparrow$ & $C \downarrow$ & $D \uparrow$ & $S \uparrow$ & $C \downarrow$ \\
& (m) & (\%) & (\%) & (m) & (\%) & (\%) & (m) & (\%) & (\%) & (m) & (\%) & (\%) \\
\midrule

Ours            & \textbf{9.67} & \textbf{97.00} & \textbf{2.37}  
                & \textbf{9.51} & \textbf{92.60} & \textbf{8.78}  
                & \textbf{9.29} & \textbf{87.40} & \textbf{13.45} 
                & \textbf{8.94} & \textbf{82.20} & \textbf{24.33} \\
                
\midrule

Blind           & 5.53 & 28.40 & 51.11 
                & 5.38 & 23.20 & 75.56 
                & 5.08 & 15.60 & 86.67 
                & 4.62 & 9.40  & 93.33 \\
Ours w/o Virtual Obstacle Stage 
                & 6.19 & 51.60 & 13.18 
                & 5.88 & 45.80 & 20.88 
                & 5.42 & 39.40 & 29.67 
                & 5.06 & 30.80 & 36.15 \\
Ours w/o ReLU Clearance Penalty      
                & 8.22 & 75.60 & 16.48 
                & 7.71 & 62.80 & 23.07 
                & 7.25 & 53.60 & 31.87 
                & 6.97 & 46.20 & 39.56 \\
Ours w/o Semantic Map      
                & 9.55 & \textbf{97.00} & 7.46  
                & 9.36 & 92.20 & 11.28 
                & 8.92 & 84.20 & 17.33 
                & 8.59 & 77.80 & 26.22 \\
\bottomrule
\end{tabular}
}
\vspace{-10pt}
\end{table*}

\subsection{Experimental Setup}
We compare {\pname} with baselines and ablations:

\begin{itemize}
    \item \textbf{Blind:} A blind policy with standard walking gait.
    \item \textbf{Ours w/o Virtual Obstacle Stage:} An ablation of our policy that removes the first stage.
     \item \textbf{Ours w/o ReLU Clearance Penalty:} This policy uses a strict trajectory tracking constraint $-(z - s_{\text{feet}} \cdot {\sin(\pi \phi_{\text{swing}}^{i})})^2$ to replace the ReLU Clearance Penalty.
    \item \textbf{Ours w/o Semantic Map:} Search using high-level information instead of semantic information.
    \item \textbf{Ours (full):} Our policy trained via a two-stage RL approach with semantic input and ReLU clearance penalty. 
\end{itemize}

Experiments are conducted in Isaac Sim. 
For the simulation evaluation, we benchmark each policy on a $10\text{m} \times 2\text{m}$ straight-line track populated with varying densities of obstacle. The robot is commanded to traverse the track with a target forward velocity of $0.4\text{ m/s}$.

We evaluate the performance using three primary metrics:
\begin{itemize}
    \item \textbf{Success Rate ($S$, \%)}: The percentage of trials in which the robot successfully completes the obstacle course.
    \item \textbf{Average Distance to Failure ($D$, m)}: The average distance traveled by the robot before encountering a failure. (This metric is selected to mitigate evaluation biases caused by bimodal performance distributions.)
    \item \textbf{Step Collision Rate ($C$, \%)}: The percentage of footsteps that collide with obstacles relative to the total number of steps taken during the traversal.
\end{itemize}

\subsection{Simulation Evaluation}
\subsubsection{Quantitative Analyses}

\begin{figure*}[t]   
    \centering    
    \includegraphics[width=0.95\linewidth]{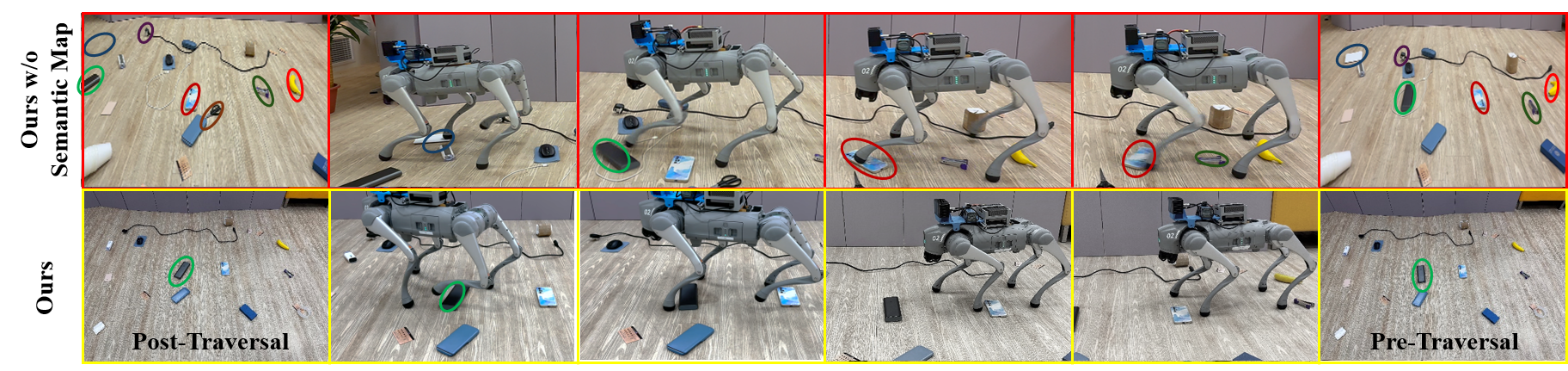}   
    \caption{Qualitative comparison of obstacle avoidance in a real cluttered environment.We compare our semantic-aware locomotion policy (Bottom) against an elevation-only baseline (Top). Elevation maps completely fail to capture small, low-profile objects (Fig~\ref{fig:overview}). Consequently, Ours w/o Semantic Map degrades into a nearly blind policy, failing to adjust foot placements to avoid obstacles. As highlighted by the colored circles, the robot frequently steps on or kicks the scattered objects; the resulting environmental disruption is highly evident when comparing the Pre-Traversal and Post-Traversal states (the first and last columns). In contrast, our full framework leverages explicit semantic cues to achieve precise obstacle avoidance. The environment remains almost entirely undisturbed, with the exception of a negligible displacement of a power bank caused by an incidental grazing contact during the swing leg's descent.}
    \label{fig:baseline_experiments}
\end{figure*}

The quantitative results of our experiments are summarized in Table~\ref{tab:quantitative_results}. To ensure a fair and statistically robust comparison, all policies are evaluated across an identical set of 500 randomly generated environments for each density configuration. The reported quantitative metrics are averaged over these 500 independent trials. In general, {\pname} \ consistently outperforms all baseline methods across different obstacle densities, particularly in highly dense environments.

\textbf{Traversal Capability and Survival:} 
As expected, the Blind policy finds it almost impossible to navigate entire obstructed terrains, resulting in the lowest average distance to failure ($D$) and the lowest success rate ($S$). In contrast, Ours (full) maintains a near-perfect $D$ of approximately 9~m, demonstrating robust locomotion and sustained survival even when faced with heavily cluttered obstacles. 

\textbf{Obstacle Avoidance Performance:} 
The effectiveness of our two-stage RL approach is most evident in the step collision rate ($C$). Ours (full) achieves the lowest $C$ among all methods, effectively reducing unintended foot-obstacle collisions by 74\% to 95\% (depending on obstacle density) relative to the blind baseline. Conversely, the Blind policy possesses almost no capability for obstacle avoidance; as environmental density increases, almost every footstep inevitably results in a collision with the terrain debris.
\subsubsection{Detailed Ablation Analysis}
We perform a detailed ablation analysis focusing on the two-stage RL curriculum, the reward formulation, and the exteroceptive modalities, as detailed in the Table~\ref{tab:quantitative_results}.

\textbf{Effectiveness of the Two-Stage Curriculum:} Removing the initial virtual obstacle stage (Ours w/o Virtual Obstacle Stage) leads to catastrophic performance degradation. For instance, at a moderate density of 15 obstacles/$\mathrm{m}^{2}$, the success rate ($S$) plummets from 92.60\% to 45.80\%, and the mean distance to failure ($D$) drops to 5.88~m. Without the soft dynamic constraints provided in the first stage, it is difficult for the agent to learn stable obstacle avoidance and stable gait from the physical feedback with obstacles during its early exploration phase. This highlights that our two-stage RL is indispensable for resolving the fundamental conflict between spatial exploration and strict temporal gait adherence.

\textbf{Flexible Foot Clearance vs. Strict Tracking:} When substituting our unilateral ReLU-based clearance penalty with a strict bilateral trajectory tracking constraint (Ours w/o ReLU Clearance Penalty), we observe a significant deterioration in obstacle avoidance. At 15 obstacles/$\mathrm{m}^{2}$, the step collision rate ($C$) increases by a factor of 2.6, rising from 8.78\% to 23.07\%. A strict trajectory constraint forces the foot to rigidly adhere to a predefined mathematical curve, depriving the agent of the spatial freedom required to dynamically elevate its feet higher than the nominal trajectory when negotiating unpredictable debris. The ReLU-based penalty, on the contrary, enforces only a minimum safe clearance, granting the policy the necessary flexibility to execute collision-free swing trajectories.

\textbf{Semantic Awareness vs. Geometric Proxy:} Interestingly, the policy stripped of semantic inputs (Ours w/o Semantic Map) maintains a Success Rate highly comparable to our full pipeline across the physical clutter track (e.g., both achieve 97.00\% at 10 obstacles/$\mathrm{m}^{2}$). However, it exhibits a noticeably higher Step Collision Rate (e.g., 7.46\% vs. 2.37\% at 10 obstacles/m²). This behavior is fundamentally expected in simulation. Since the scattered obstacles inherently possess prominent physical heights, the neural network efficiently learns to exploit extreme elevation gradients from the geometric map as a proxy for unsafe footholds. However, we explicitly retain this baseline to emphasize the conceptual ceiling of elevation-only methods. While elevation serves as an effective obstacle proxy on flat training terrains, it fails under real-world complexities. First, on uneven topographies like stairs, pure geometry becomes ambiguous, as height variation may indicate a safe foothold. Second, small obstacles are easily obscured by real sensor noise, underscoring the necessity of explicit semantic information.
\subsection{Sim-to-Real Experiments}
To validate the practical viability of {\pname}, we conduct qualitative real-world tests in an unstructured indoor environment. The test track is deliberately populated with fragile and low-profile everyday clutter, including power cables, smartphones, boxes, and small stationery.

\textbf{Sensor Noise and Map Modality Comparison:} These small, low-profile objects represent the exact edge cases where traditional geometric proxies fail. In our visualization, the physical depth sensor's inherent noise floor completely obscures the millimeter-level height variations of these items, rendering the elevation map virtually indistinguishable from the flat ground. In stark contrast, the semantic map distinctly highlights these objects as high-cost hazards. This visual validation confirms the absolute necessity of semantic-geometry disentanglement for reliable real-world perception.

\textbf{Semantic Avoidance:} The core intelligence of {\pname} \ lies not merely in elevating the swing leg, but in its proactive foothold judgment. During experiment, if our policy predicts a foothold landing directly on a semantically hazardous object (e.g., a smartphone), the policy dynamically intervenes. It actively shortens the stride or step off to the side to secure a safe foothold immediately before the object. Subsequently, on the next gait cycle, it commands a high-clearance swing trajectory to cleanly step over the hazard. While a traditional blind controller might occasionally step over an object by pure luck if its foot happens to land perfectly in front of it, it fundamentally lacks this spatial judgment and proactive stride modulation.

\textbf{Baseline Comparison:} We qualitatively compare {\pname} \ against a \textbf{Ours w/o Semantic Map} baseline. The baseline policy exhibits a severe "bulldozer" behavior: entirely oblivious to the surrounding fragility, it marches forward rigidly. Consequently, it collides with, kicks away, and aggressively tramples the scattered debris, posing a critical danger to both the robot and the objects. Conversely, {\pname} \ demonstrates significantly more deliberate navigation, effectively mitigating the majority of severe collisions with fragile debris. This highlights the potential of semantic-aware locomotion to improve operational safety in unstructured, real-world environments.

\section{Conclusion}
In this work, we introduced {\pname}, a semantic-aware locomotion framework that achieves precise foothold selection in dense, unstructured environments. By explicitly disentangling environmental semantics from geometric heights and employing a two-stage reinforcement learning curriculum, {\pname} successfully overcomes the geometric ambiguities and perceptual failure modes inherent to traditional elevation proxies. Extensive evaluations across both simulation and real-world indoor scenarios confirm that our approach significantly mitigates step collision rates.

\textbf{Limitation and Future Work:}
While the proposed {\pname} \ framework significantly enhances semantic-aware locomotion, simulation and deployment reveal limitations in both rigid-body dynamics and semantic representation. In terms of dynamics and hardware, extreme asymmetric footholds induce uncompensated angular momentum, occasionally resulting in observable yaw drift or pushing the joints near kinematic singularities. Additionally, it is extremely difficult for the real robots to replicate the foot trajectory and footswing height as seen in simulations. A noticeable sim-to-real gap stems from the semantic simulation environment. To manage computational overhead during training, complex real-world objects are approximated using simplified geometric primitives (e.g., cubes). This approximation limits the policy's exposure to realistic object boundaries. Furthermore, our current pipeline compresses environmental semantics into a "fragility" cost. However, real-world applications encompass open-vocabulary semantics that dictate diverse interactive behaviors. Future work will explore integrating high-fidelity semantic simulators and Vision-Language-Action (VLA) models to achieve comprehensive open-vocabulary semantic comprehension and generalized locomotion.


\ifCLASSOPTIONcaptionsoff
  \newpage
\fi

\bibliographystyle{IEEEtran}
\bibliography{IEEEexample}

@article{hwangbo2019agile,
  title   = {Learning Agile and Dynamic Motor Skills for Legged Robots},
  author  = {Hwangbo, Jemin and Lee, Joonho and Dosovitskiy, Alexey and Bellicoso, Dario and Tsounis, Vassilios and Koltun, Vladlen and Hutter, Marco},
  journal = {Science Robotics},
  volume  = {4},
  number  = {26},
  pages   = {eaau5872},
  year    = {2019},
  doi     = {10.1126/scirobotics.aau5872}
}

@inproceedings{roth2024viplanner,
  title={Viplanner: Visual semantic imperative learning for local navigation},
  author={Roth, Pascal and Nubert, Julian and Yang, Fan and Mittal, Mayank and Hutter, Marco},
  booktitle = {Proc. IEEE Int. Conf. Robot. Autom. (ICRA)},
  year={2024}
}

@article{miki2022perceptive,
  title   = {Learning Robust Perceptive Locomotion for Quadrupedal Robots in the Wild},
  author  = {Miki, Takahiro and Lee, Joonho and Hwangbo, Jemin and Wellhausen, Lorenz and Koltun, Vladlen and Hutter, Marco},
  journal = {Science Robotics},
  volume  = {7},
  number  = {62},
  pages   = {eabk2822},
  year    = {2022},
  doi     = {10.1126/scirobotics.abk2822}
}

@inproceedings{kumar2021rma,
  title     = {RMA: Rapid Motor Adaptation for Legged Robots},
  author    = {Kumar, Ashish and Fu, Zipeng and Pathak, Deepak and Malik, Jitendra},
  booktitle = {Robotics: Science and Systems (RSS)},
  year      = {2021},
  doi       = {10.15607/RSS.2021.XVII.011},
  eprint    = {2107.04034},
  archivePrefix = {arXiv},
  primaryClass   = {cs.RO}
}

@article{fankhauser2018terrain,
  title   = {Probabilistic Terrain Mapping for Mobile Robots with Uncertain Localization},
  author  = {Fankhauser, P{\'e}ter and Bloesch, Michael and Hutter, Marco},
journal = {IEEE Robot. Autom. Lett.},
  year    = {2018}
}

@inproceedings{rudin2022advancedskills,
  title     = {Advanced Skills by Learning Locomotion and Local Navigation End-to-End},
  author    = {Rudin, Nikita and Hoeller, David and Bjelonic, Marko and Hutter, Marco},
booktitle = {Proc. IEEE/RSJ Int. Conf. Intell. Robots Syst. (IROS)},
  year      = {2022},
  doi       = {10.1109/IROS47612.2022.9981198}
}

@inproceedings{yang2023corl_semanticsloco,
  author       = {Yang, Yuxiang and Meng, Xiangyun and Yu, Wenhao and Zhang, Tingnan and Tan, Jie and Boots, Byron},
  title        = {Learning Semantics-Aware Locomotion Skills from Human Demonstration},
  booktitle    = {Proceedings of The 6th Conference on Robot Learning (CoRL)},
  series       = {Proceedings of Machine Learning Research},
  volume       = {205},
  pages        = {2205--2214},
  year         = {2023},
  url          = {https://proceedings.mlr.press/v205/}
}

@inproceedings{aegidius2025watch,
  title={Watch your stepp: Semantic traversability estimation using pose projected features},
  author={{\AE}gidius, Sebastian and Hadjivelichkov, Dennis and Jiao, Jianhao and Embley-Riches, Jonathan and Kanoulas, Dimitrios},
  booktitle={2025 IEEE International Conference on Robotics and Automation (ICRA)},
  pages={2376--2382},
  year={2025},
  organization={IEEE}
}

@article{fan2022riskaware,
  author       = {Fan, David D. and Dey, Sharmita and Agha-mohammadi, Ali-akbar and Theodorou, Evangelos A.},
  title        = {Learning Risk-aware Costmaps for Traversability in Challenging Environments},
journal = {IEEE Robot. Autom. Lett.},
  year         = {2022}
}

@article{miles2023terrainaware,
  author       = {Miles, Michael J. and Biggie, Harel and Heckman, Christoffer},
  title        = {Terrain-aware semantic mapping for cooperative subterranean exploration},
  journal      = {Frontiers in Robotics and AI},
  volume       = {10},
  pages        = {1249586},
  year         = {2023},
  doi          = {10.3389/frobt.2023.1249586}
}

@article{yue2025safetypath,
  author       = {Yue, Rui and Feng, Lifeng and Ma, Lei and Zhang, Muhua and Shen, Kai and Sun, Yongkui},
  title        = {Safety Path Planning for Quadruped Robots Optimized by Multi-Sensor Fusion},
  journal      = {IFAC-PapersOnLine},
  volume       = {59},
  number       = {27},
  pages        = {55--60},
  year         = {2025},
  doi          = {10.1016/j.ifacol.2025.12.078}
}

@article{ha2025learning,
  title={Learning-based legged locomotion: State of the art and future perspectives},
  author={Ha, Sehoon and Lee, Joonho and van de Panne, Michiel and Xie, Zhaoming and Yu, Wenhao and Khadiv, Majid},
  journal={The International Journal of Robotics Research},
  volume={44},
  number={8},
  pages={1396--1427},
  year={2025},
  publisher={Sage Publications Sage UK: London, England}
}

@article{grandia2023perceptive,
  title={Perceptive locomotion through nonlinear model-predictive control},
  author={Grandia, Ruben and Jenelten, Fabian and Yang, Shaohui and Farshidian, Farbod and Hutter, Marco},
  journal={IEEE Transactions on Robotics},
  volume={39},
  number={5},
  pages={3402--3421},
  year={2023},
  publisher={IEEE}
}

@article{kim2019highly,
  title={Highly dynamic quadruped locomotion via whole-body impulse control and model predictive control},
  author={Kim, Donghyun and Di Carlo, Jared and Katz, Benjamin and Bledt, Gerardo and Kim, Sangbae},
  journal={arXiv preprint arXiv:1909.06586},
  year={2019}
}

@INPROCEEDINGS{9981198,
  author={Rudin, Nikita and Hoeller, David and Bjelonic, Marko and Hutter, Marco},
booktitle = {Proc. IEEE/RSJ Int. Conf. Intell. Robots Syst. (IROS)},
  title={Advanced Skills by Learning Locomotion and Local Navigation End-to-End}, 
  year={2022}
  }

@article{jenelten2020perceptive,
  title={Perceptive locomotion in rough terrain--online foothold optimization},
  author={Jenelten, Fabian and Miki, Takahiro and Vijayan, Aravind E and Bjelonic, Marko and Hutter, Marco},
journal = {IEEE Robot. Autom. Lett.},
  year={2020}
}

@article{kim2024learning,
  title={Learning semantic traversability with egocentric video and automated annotation strategy},
  author={Kim, Yunho and Lee, Jeong Hyun and Lee, Choongin and Mun, Juhyeok and Youm, Donghoon and Park, Jeongsoo and Hwangbo, Jemin},
journal = {IEEE Robot. Autom. Lett.},
  year={2024}
}

@inproceedings{cai2022risk,
  title={Risk-aware off-road navigation via a learned speed distribution map},
  author={Cai, Xiaoyi and Everett, Michael and Fink, Jonathan and How, Jonathan P},
booktitle = {Proc. IEEE/RSJ Int. Conf. Intell. Robots Syst. (IROS)},
  year={2022}
}

@inproceedings{erni2023mem,
  title={MEM: Multi-modal elevation mapping for robotics and learning},
  author={Erni, Gian and Frey, Jonas and Miki, Takahiro and Mattamala, Matias and Hutter, Marco},
booktitle = {Proc. IEEE/RSJ Int. Conf. Intell. Robots Syst. (IROS)},
  year={2023}
}

@inproceedings{achat2022path,
  title={Path planning incorporating semantic information for autonomous robot navigation},
  author={Achat, Silya and Marzat, Julien and Moras, Julien},
  booktitle={19th International Conference on Informatics in Control, Automation and Robotics (ICINCO) 2022},
  pages={285--295},
  year={2022},
  organization={SCITEPRESS-Science and Technology Publications}
}

@inproceedings{maturana2017real,
  title={Real-time semantic mapping for autonomous off-road navigation},
  author={Maturana, Daniel and Chou, Po-Wei and Uenoyama, Masashi and Scherer, Sebastian},
  booktitle={Field and Service Robotics: Results of the 11th International Conference},
  pages={335--350},
  year={2017},
  organization={Springer}
}

@article{kremer2023s,
  title={S-nav: Semantic-geometric planning for mobile robots},
  author={Kremer, Paul and Bavle, Hriday and Sanchez-Lopez, Jose Luis and Voos, Holger},
  journal={arXiv preprint arXiv:2307.01613},
  year={2023}
}

@inproceedings{margolis2023walk,
  title={Walk these ways: Tuning robot control for generalization with multiplicity of behavior},
  author={Margolis, Gabriel B and Agrawal, Pulkit},
  booktitle = {Conf. Robot Learn. (CoRL)},
  year={2023}
}

@inproceedings{siekmann2021sim,
  title={Sim-to-real learning of all common bipedal gaits via periodic reward composition},
  author={Siekmann, Jonah and Godse, Yesh and Fern, Alan and Hurst, Jonathan},
  booktitle = {Proc. IEEE Int. Conf. Robot. Autom. (ICRA)},
  year={2021}
}

@book{raibert1986legged,
  title={Legged robots that balance},
  author={Raibert, Marc H},
  year={1986},
  publisher={MIT press}
}

@article{ji2022concurrent,
  title={Concurrent training of a control policy and a state estimator for dynamic and robust legged locomotion},
  author={Ji, Gwanghyeon and Mun, Juhyeok and Kim, Hyeongjun and Hwangbo, Jemin},
  journal = {IEEE Robot. Autom. Lett.},
  year={2022}
}

@article{margolis2024rapid,
  title={Rapid locomotion via reinforcement learning},
  author={Margolis, Gabriel B and Yang, Ge and Paigwar, Kartik and Chen, Tao and Agrawal, Pulkit},
journal = {Int. J. Robot. Res.},
  year={2024},
  publisher={SAGE Publications Sage UK: London, England}
}

@article{sun2025yolo,
  title={YOLO-E: a lightweight object detection algorithm for military targets},
  author={Sun, Yong and Wang, Jianzhong and You, Yu and Yu, Zibo and Bian, Shaobo and Wang, Endi and Wu, Weichao},
  journal={Signal, Image and Video Processing},
  volume={19},
  number={3},
  pages={241},
  year={2025},
  publisher={Springer}
}

@article{mittal2025isaac,
  title={Isaac lab: A gpu-accelerated simulation framework for multi-modal robot learning},
  author={Mittal, Mayank and Roth, Pascal and Tigue, James and Richard, Antoine and Zhang, Octi and Du, Peter and Serrano-Munoz, Antonio and Yao, Xinjie and Zurbr{\"u}gg, Ren{\'e} and Rudin, Nikita and others},
  journal={arXiv preprint arXiv:2511.04831},
  year={2025}
}

@article{schulman2017proximal,
  title={Proximal policy optimization algorithms},
  author={Schulman, John and Wolski, Filip and Dhariwal, Prafulla and Radford, Alec and Klimov, Oleg},
  journal={arXiv preprint arXiv:1707.06347},
  year={2017}
}

@IEEEtranBSTCTL{IEEEexample:BSTcontrol,
  CTLuse_forced_etal       = "yes",
  CTLmax_names_forced_etal = "3",
  CTLnames_show_etal       = "2"
}

\end{document}